# A FRAMEWORK: CLUSTER DETECTION AND MULTIDIMENSIONAL VISUALIZATION OF AUTOMATED DATA MINING USING INTELLIGENT AGENTS

R. Jayabrabu[1], Dr. V. Saravanan[2], Prof. K. Vivekanandan[3]

[1]PhD Research Scholar, Department of Computer Applications, Bharathiar University
[2]Professor & Director, Department of Computer Applications, Sri Venkateswara College of Computer Applications and Management
[3]Professor, School of Management, Bharathiar University
[1]jayabrabu@gmail.com, [2]tvsaran@gmail.com, [3]vivekbsmed@gmail.com

*ABSTRACT:*
*Data Mining techniques plays a vital role like extraction of required knowledge, finding unsuspected information to make strategic decision in a novel way which in term understandable by domain experts. A generalized frame work is proposed by considering non – domain experts during mining process for better understanding, making better decision and better finding new patters in case of selecting suitable data mining techniques based on the user profile by means of intelligent agents.*

## 1. INTRODUCTION

Past decades has seen a dramatic increase in the amount of information of data being stored in different electronic format. The problem is how to maintain these data and also extraction of required knowledge. Data mining techniques can be viewed as a result of the natural evaluation of information technology [4] [16]. The process of analyzing the observational data sets to find unsuspected relationships and to summarize the data in novel ways those are both understandable and useful to the data user called as data mining.[16] [12] It is the achievement of relevant knowledge that can allow you to make strategic decisions which will allow you to proceed further. The techniques used in data mining are more difficult to understand by general user [1]. A system which performs the process autonomously needs more user interaction by way of selecting the appropriate attribute. Automated data mining, minimizes this problem by selecting the attributes based on the user specified objective. It also chooses the best data mining technique for knowledge extraction [9]. The level of automation incorporated in this mining system is an important issue. Detecting the interesting patterns and finding out the appropriate knowledge from the output is also an interesting issue in data mining. This research work addresses this issue with the help of software agents. [7] The software agents help in detecting the clusters automatically. It also assists in viewing the cluster results graphically. Multidimensional visualization is used to view the results in a more meaningful way.

## 2. OBJECTIVE

To develop a user friendly data mining system using intelligent agent through automated approach is the objective function of this research work. It mainly deals about, developing a neutral system by using well known data mining techniques. Intelligent agents are also used in the above mentioned techniques to implement the automated process [1]. The developed automated system gets the input values form the user and chooses the appropriate data mining techniques with required parameters by using intelligent agents [7] [3]. The system also detects the quality of clusters with respect to user profile. Finally, the system displays the newly gained knowledge after clustering by choosing appropriate visualization technique. The user agent is used to navigate user activates [4]. While user using system frequently, the previous history or activities of the particular user will be prompted once the particular user logged in the system after some time. The data mining agent is used to perform different types of data analysis by selecting different mining algorithms based on the user navigation for better results in a shorter time [5]. Intelligent agents are also used to select the required parameter for the given problem domain within the autonomous system for better visualization [13] [14].

## 3. RELATED WORK

Berry and Linoff [4] proposed automated data mining by taking a picture with an automatic camera. The quality of results is not always as good as what can be achieved by an expert; however the ease of use empowers non- expert users to achieve reasonable results with minimum effort. This method is considered in this research work to automate the selection process of user required attributes and mining algorithms for better decision making process. By considering this approach, the time taken to select the attributes and mining algorithm is reduced while compared with manually process is being considered.

Saravanan and Vivekanandan[35] proposed an automated data mining system which encompasses familiar data mining algorithms. According to author the system will automatically select the appropriate data mining technique and select the necessary field needed from the database at the appropriate time without expecting the users to specify the specific techniques and the parameters. Association and Classification rule mining is incorporated in this approach with the help of software agents. The system also has multiple association and classification techniques and selects the appropriate techniques based on user interest / data type / data size. Automatic detection of clusters and multi-dimensional visualization is not being considered by this author.

Ayse Yasemin SEYDIM's [1] explained more on agents, the special types of software applications, has become a very popular paradigm in computing in recent years. The author states that, the agent based studies can be implemented for clustering, classification, and summarization. Some of the reasons where agents are more flexibility, modularity and general applicability to a wide range of problems. Recent increase in agent-based applications is also because of the technological developments in distributed computing, robotics and the emergence of object-oriented programming paradigms. Advances in distributed computing technologies have given rise to use of agents that can model distributed problem solving.

Vuda Sreenivasa Rao [40] explained communications among the agents with in multi-agent system. According to the author, multi-agent system often deals with complex applications that required to solve the existing problem during data mining process in distributed system with individual and collective behaviors of the agents depends on the observed data from distributed system. Based on this concept, an integration of multi-agent system with data mining is incorporated and it also defines how multiple agents are communicated with respect to specific applications. Declaration of different agents with respect to specific task and communication behavior among agents is considered in this research work to meet the user requirements.

Marcos M. Campos**,** Peter J. Stengard and Boriana L. Milenova [22] proposes a new approach to the design of data mining applications platform to targeted user communities. This approach uses a data-centric focus where information are stored in a location and by implementing automated methodologies to make data mining process more accessible to non-experts. The automated methodologies are exposed through high-level interfaces. This frame work hides the data mining concepts away from the users, helping to bridge the conceptual gap generally associated with data mining. Automated mining algorithm used in this approach is classification and regression techniques. Clustering techniques is being used in this research work with the help of automated data mining system to create new cluster for user community.

Faraz Zaidi, Daniel Archambault and Guy Melancon [FDG, 41] proposed a metric to evaluate the quality of a cluster based on the path length of the elements of a cluster. Generally, metrics based on density and cut size prove to be adequate for networks having densely connected nodes or cliques. The author proposed a very simple and intuitive metric in which instead of considering density and cut size as fundamental component to evaluate the quality of the clustering algorithm, they suggested the mutuality and compactness of a cluster that can also be easily evaluated using a single quantities measure: the average path length between all the nodes of a cluster. The average path length determines the closeness of the elements of a cluster which represent how far apart any two nodes lie to each other. The author concentrated only on internal quality metrics. As a part of future work, an extended study on algorithms that can generate a benchmark cluster for given data sets with varying properties in multidimensional dataset is proposed and it is addressed with some extent in this research work.

David Meunier and Helene Paugam-Moisy [DH, 42] focused on Girvan & Newman (GN) method which was created by Girvan & Newman to determine the clusters in a given undirected graph, without any restriction on the data size and number of clusters. GN method is based two-pass algorithm where the first pass removes the edges with highest edge – betweenness centrality and cluster building in the second pass. Author proposed an another method based on GN method instead of undirected graph to directed graph in the name of arc-linked cluster detection method with neural networks concepts to achieves a narrower and higher modularity peak than GN method. The resultant value is more pertinent optimal set of clusters. The time complexity of Arc-linked based is higher than GN method. The concept discussed by these authors is considered in this research work during cluster detection phases after cluster formation to detect the quality of cluster by means of attribute selection from the given database.

Eleni Mangina [9] discusses real world monitoring engineering applications and specifies the role of knowledge engineers in complexity and diversity of tasks associated with specific problem domain. This paper also deals about tackling simultaneously different types of knowledge (inaccurate, incorrect or redundant) from different data sources that require being processed using different reasoning mechanisms. Within this paper, an intelligent agent-based platform is being considered for implementation, where the approach of integrating the use of two or more techniques is taken, in order to combine their different strengths and overcome each other's weaknesses and generate hybrid solutions. Different types of knowledge are discovered using different data mining techniques based on the user is considered to implement in this research.

Wout Dullaert, Tijs Neutens, Greet Vanden Berghe [WTG, 43] implements an intelligent agent-based communication for particular platform. Intelligent agents are used in the form of high potential output such as increase cost efficiency, better service and safety communication among the agents. They are also autonomous, communicative and intelligent. Author also proposed real-time decision is also possible with the presence of intelligent agent. Agents are used to overcome the quality, reliable service, trust concerns and confidentiality during the exchange process. Agent technology is used for automated transport process. In this research work, intelligent agent-based concept is considered for cluster formation and cluster visualizations.

Nigel Robinson, Mary Shapcott [26] proposed a visualization aids (beyond charts and graphs) by consider virtual data mining environment and data set as liquid data. The author focused was based on limitations of the existing visualization methods and problems faced by the user while visualization. As innovation, the modified visualization result is in the form of 3D game representation so that all users can easy to understand without having domain knowledge. The minimization of user difficulties is taken into account in our research process during visualization. In this paper, automation of the process is not yet considered for result oriented. Because of this, it is difficult to find the results by rare user. From the above procedure, concept of visualization is taken into account as a part of this research work in which an automated data mining system is implemented to sense the user behavior for visualization so as to minimize the difficulties faced by the user during graphical representation of the results.

Dr. Ping Chen, Dr. Chenyi Hu, Dr. Heloise Lynn, and Yves Simon [28] discuss about visualization of high dimensional data which is very important in data analysts due to its visual nature. They also proposed a method to visualize large amount of high dimensional data in a 3-D space by dividing the high dimension data into several groups of lower dimensional data first. Then, different icons are used to represent different groups. A glyph-based technique is used to represent different set of data in the form of various color icons like line, point, polygon, etc. The visualization of high dimensional data using Glyphs – based techniques is considered in this research work for visualization of multi-dimensional data based on user expectation.

### 3.1 Need of the Approach for the Proposed Framework

To develop a user friendly data mining system using intelligent agent through automated approach is the objective function of this research work. It mainly deals about, developing a neutral system by using well known data mining techniques. Intelligent agents are also used in the above mentioned techniques to implement the automated process [35]. The developed automated system gets the input values form the user and chooses the appropriate data mining techniques with required parameters by using intelligent agents. [30]The system also detects the quality of clusters with respect to user profile. Finally, the system displays the newly gained knowledge after clustering by choosing appropriate visualization technique. The user agent is used to navigate user activates. When the user using the system frequently, the previous history or activities of the particular user will be prompted once the particular user logged in the system after some time. The data mining agent is used to perform different types of data analysis by selecting different mining algorithms based on the user navigation for better results in a shorter time. Intelligent agents are also used to select the required parameter for the given problem domain within the autonomous system for better visualization [36].

### 4. METHODOLOGY

To develop an approach which performs the process autonomously that needs more user interaction by way of selecting the appropriate user specifies objectives are the primary goal of this research [35] [36]. The developed system also chooses the best data mining technique for knowledge extraction and also for better understanding. Thus, level of automation incorporated in this system is an important issue. The automated system performs the process based on user interface agent, data mining agent and visualization agent. [26] [25] User interface agent is used to navigate the history of the user profile among the frequent user to mine the related data [27]. Data mining agent is used in this system to perform different types of data analysis by selecting different data mining algorithm based on the user profile. The profile selected by the agent is then inputted to the automated data mining systems where the clustering technique is used to find new patterns. [32] Various clustering techniques such as k-means, CLARA, CLARNS, k- prototype, k-mode, etc is being used to perform cluster analysis[31] [34]. The automated data mining system chooses any one of the mining algorithm based upon the data types, size of the data, and user profile. Attribute selection (prioritization, ranking) techniques are used to select the appropriate attributed to complete the mining process by means of intelligent agent. The intelligent agents used in this automated system, [21] analyses the quality of cluster among the clusters produced by different data mining techniques with respect to the history of profile[17] [18]. In this automated system, agent selects the appropriate visualization tool (1D (raw format), 2D (x-axis, y-axis) and 3D(x, y, z-axis)) for better decision making [26] [28] [32]. From these aspects, a framework is developed to analyze the cluster, detect the cluster and visualization by means of automated system and intelligent agent technology [25]. The methodology adopted in this research work is given below.

**Proposed Methodology**

*Step 1: Start*

*Step 2: Navigation of User Profile.*

*Step 3: Design of a suitable procedure for the intelligent agents to Interact with User Profile and Clustering Techniques.*

*Step 4: Selection of appropriate Clustering Techniques.*

*Algorithm Selection.*

*Attributes Selection. (Priority, Ranking)*

*Step 5: Design a suitable procedure for cluster detection*

*Step 5: Design a suitable procedure for the Software Agent to select the appropriate Visualization tool.*

*Step 6: Multi Dimensional Visualization of Results.*

*Step 7:Stop*

*Figure4a. Procedure for Proposed Methods*

# 5.Intelligent Agent

Artificial intelligence and agent systems have been closely inter-linked with each other over few decades [15] [17] [6] [37]. The artificial intelligence is the study of components intelligence, but agent deals with integration of those components. Thus, agent is defined as an entity that vaguely a software using techniques from artificial intelligence to assist a human user of a specific application [11] [30] [8] [27] [33]. According to the definition of agents have the following properties:

1. Agents are autonomous in nature: operates without direct control with user and other agents.
2. Agents are also called as social communicator among the agents.
3. Agents by itself move towards its defined goal.

## 5.1 User Interface Agent

The user interface agent interacts with the user in assisting him / her to perform data analysis and data mining activities [35] [10] [15] [24]. The user can provide a general description of the problem at hand in terms of high level goals and objectives, or provide specific details about the data analysis or mining task to be performed.

International Journal of Artificial Intelligence & Applications (IJAIA), Vol.3, No.1, January 2012

**Developed Procedure for User Interface Agent**

*Step 1: Start*
*Step 2: Input the user objectives*
*Step 3: Check user history*
*Step 4: If new users go to step 6*
*Step 5: If user already exists*

  *5.1 Retrieve the possible interested patterns.*
  *5.2 Wait for the user response*
  *5.3 Capture user navigation and store it in user history*
  *5.4 Go to step 7*

*Step 6: Retrieve similar interested patterns,*
        *Capture user navigation and store it in*
        *user history*
*Step 7: Transfer control to "Data mining Agent"*
*Step 8: End*

*Figure 5.1a: Procedure for User Interface Agent*

## 5.2 Ranking Agent

Once the user preference is identified the ranking agent helps in ranking the attributes. Ranking is done by number of factors such as query weight, memory space, properties of the attribute type and similarity between the objects [29]. The highest ranked attributes are taken for clustering.

**Developed Procedure for Ranking Agent**

*Step 1: Start*
*Step2: Calculate the scoring function for each attribute*
      *Step 2.1: Classify the types of*
                *attributes*
      *Step 2.2: Assign Score Value*
      *for each attributes*
*Step 3: Calculate Query Weight*
      *Step 3.1: Usage of attributes*
      *Step3.2:Assigning of weightage*
*Step4:Identify the relationship and importance of the*
       *ranked attributes*
*Step 5: End*

*Figure 5.2 a: Procedure for Ranking Agent*

## 5.3 Data Mining Agent

The data-mining agent implements specific data mining methods and algorithms. The data mining agent contains specific clustering algorithms implemented in house, that may have been installed within the environment [2] [23][36]. Thus, the data mining agent is responsible for performing the actual data mining activity and generating the results.

**Developed Procedure for Data Mining Agent**

*Step 1: Start*
*Step 2: Identify and Select the ranked attributes (by calling Ranking agents)*
*Step 3: Categorize the appropriate clustering algorithm suitable for the attributes*
*Step 4: Cluster data based on the suitable algorithms, k-means, k- medoids, k-modes, CLARA and CLARANS.*
*Step 5: Transfer the control to 'Visualization Agent'*
*Step 6: End*

*Figure 5.3a: Developed Procedure for Data Mining Agent*

## 5.4 Visualization Agent

To evaluate the effectiveness of visualization techniques, the way in which they assist and complement the data mining process need to be understood. [23][36]The relationship between data mining and visualization process can be explained with few conditions visualization of data set can be defined as combination of various methods or user priority to a approach of selecting and indicating what patterns should be displayed.  It is one of the interesting techniques to establish which patters are better than the one to enhance visualization techniques. Visualization will provides more added advantage to the user for easy understanding and also increases the power of understanding the end results [19].

**Developed Procedure for Visualization Agent**

*Step 1: Start*
*Step 2: Analyze the cluster data types*
*Step 3: Choose the visualization tool based on the cluster Data*
*Step 4: Classify the attributes value based on 1D or 2D or3D*
*Step 5: Choose the appropriate visualization technique based Step4*
*Step 6: Repeat the process until Step 5*
*Step 7: Visualize the data successfully*
*Step 4: End*

*Figure 5.4a: Developed Procedure for Visualization Agent*

## 6. PROPOSED FRAME WORK FOR AUTOMATED DATA MINING USING INTELLIGENT AGENTS

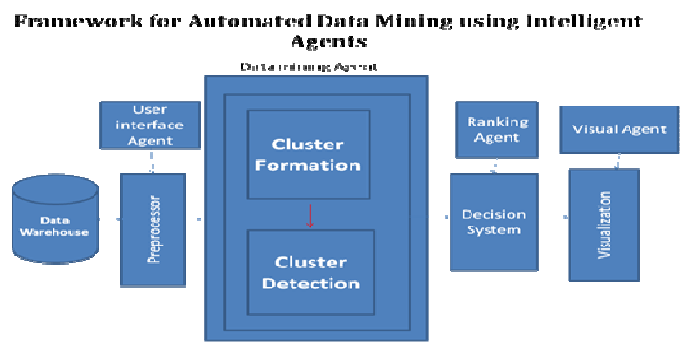

*Figure 6a: Proposed Framework for Automated Data Mining Using Intelligent Agents*

The newly developed framework is represented diagrammatically. From this framework, the automated data mining system gets the necessary data from the given database. Apart from getting data from the given database, it also gets the user information (new user or old user).  The developed system chooses the appropriate data mining techniques which are made available based on the objective functions. The algorithm considered in this developed system is partitioning algorithms. At the same time if the user already used the automated data mining system, the results are captured by the intelligent agent and stored in the database. The end result is also referred by the intelligent agent to the next user. Thus, this makes the agent technology more users friendly. The intelligent agents are used in automated process of attribute selection, ranking process, cluster formation, cluster detection and visualization.

The user interface agent is responsible for receiving user specifications and delivering results to the data mining agent. it also keeps track of user profile based on the user interactions. The data mining agent mine the result based on the parameters given by user interface agent to perform cluster formation and cluster detection. Clustering algorithms are used to formulate a new cluster, based on the user interface agent with respect to specific user profile. Then cluster detection techniques are used to detect the cluster quality for further process. The quality of cluster is identified by various parameters with help of intelligent agent. As result, the best cluster will be discovered from the known knowledge. The visual agent is used to visualize the identified cluster depends on the nature of data within the detected cluster. After completion of cluster formation and cluster detection, it transfers the result to decision system where the visual agent find suitable representation tools based on the cluster nature. Finally, the end result is visualized in terms of 1D, 2D or 3D by visual agent. Thus, the entire process is monitored as well as executed by automated intelligent system based on the user profile. This makes the less domain knowledge users more convenient and understandability.

In this research work different types of agents are used to perform the operations on behalf of user so that, the data mining result will be productive and knowledgeable for less domain knowledge user. Agents used in this framework are for reliable communication, cooperation among the agents, and finally coordination among the other agents within the system to perform some specific tasks. The overall processes are carried out by different agents with in the automated intelligent system.

## 6.1 User Interface Agent

The automated system is executed by navigating and analyzing the user profile [38]. Based on the user behavior with the data mining system and also by find the user profile (new or existing). After finding the nature of the user profile, the productive attributes are mined from user agent by data mining agent where the actual mining process takes place. The identification of users are based on the selection of data set like number of records, number of attributes, data types, etc., User interface agent analyses all the information with respect to the user behaviors and transfer the control to data mining agent for further process.

## 6.2 Data Mining Agent

Data mining agent has two major roles. They are cluster formation and cluster detection based on the parameter of user profile agent.

### 6.2a Cluster Formations

Cluster formation is defined as grouping of objects that are similar to one another within the same cluster and are dissimilar to the objects with other clusters. Partitioning approach is considered to cluster different types of attributes, numeric and categorical data. Clustering are also done depends upon data set, data size, and data types. Data mining agent will choose the appropriate clustering algorithm for better cluster formations.

**6.2b Cluster Detections**

Data mining agent is used here to analyze the formulated clusters quality based on quality parameters. Generally, clustering algorithms will produce clusters, based upon input data. But, all clusters are not good cluster. Intelligent agent in automated system is used to find the good clusters among various clusters.

**6.3 Visualization Agent**

The visualization agent is uses to generate various reports based on the cluster results.[39] The visualization might be 1D, 2D or 3D based on the type of cluster nature (numerical or categorical). Generally, major difficult task in data mining is report generation, that to it should be understand able by all the users. Manually it is possible to create reports but it is hard to justify the report at all times. In these aspects, an automated system is used to generate report based on the cluster nature irrespective of time and data. In this work, visualization agent will coordinate with data mining agent, based on that it will identify suitable visual method for each specific cluster.

## 7. DATA BASED CONSIDERED

Data set used in this research work is student database, which contains 50,000 records with different data types. The data types considered in this work are numerical data and categorical data. It contains nearly 40 attributes of different data types. Some of the attributes are as follows which is used in our research work are REG_NO, NAME, YEAR, SEMESTER, ASSIG_PARAMETER, MAX_MARK, MIN_MARK, LIB_ACCNO, ISSUE_DATE, RETURN_DATA, ACC_NO, SPECIALIZATION, STREET_NO, STREET_NO, PLACE, COUNTRY, MODULE_CODE, TUTOR_NAME, EMAIL_ID, TEL_PHONE, SESSION, PASS_PERCENTAGE, AVG_INT, AVG_EXT, AVG_TOT.

The main objective of this research work is to analyze student performance based some parameters depends on the user level. In our education systems, different types of evaluation techniques are emerged in the world to evaluate student performance by means of semester wise, class room wise, location wise, subject wise and gender wise, etc., All these evaluation patters and a technique is to improve the education level among student's community. The theme of this work is to analyze based on the user profile among domain experts and non - domain experts previous history not with students side. The domain expert will always starts from starch for their analysis purpose while compared to non – domain experts. The major problems identified with non-domain experts are:

      a.      Difficult to find the suitable data mining techniques

      b.      Difficult in selection appropriate attribute in the given data base

      c.      Difficult to find the interested patterns

      d.      Difficult to understand the interested pattern

      e.      Difficult in selecting an appropriate tool for visualization

The difficulties faced by the non-expert user is minimized with the help of this automated system. An automated system will perform task on behalf of users based on the previous user profile called as user history. From these, major parts of the tasks are inherited from the existing performance which was generated already by the existing users instead of creating new manual procedure with the help of domain experts. This leads to minimize the user difficulties with help of old historical data. The results along with screen shorts are show below with the help of automated system.

International Journal of Artificial Intelligence & Applications (IJAIA), Vol.3, No.1, January 2012

## 8. RESULTS AND DISCUSSIONS

The developed prototype uses the chosen data set for mining. The software agents are used to detect the clusters automatically and provides the best quality clusters. It is observed that the clusters given good knowledge and found to be useful for better decision making.

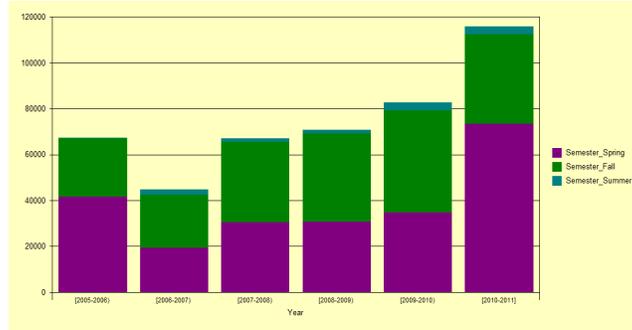

*Figure 8a: Students performance across semesters*

In figure 8a, student performance is analyzed with respect to number of students and academic year. The academic year is scheduled as three semesters namely *semester_ spring, semester_ fall* and *semester_summer* in which, student who are all studying *semester_spring* are really good in their studies when compared with other two semesters with irrespective to the students. In this scenario, the system chooses the attribute '*Semester*' and performed cluster analysis with the help of software agents. Further mining of this results are provided in figure 8b.

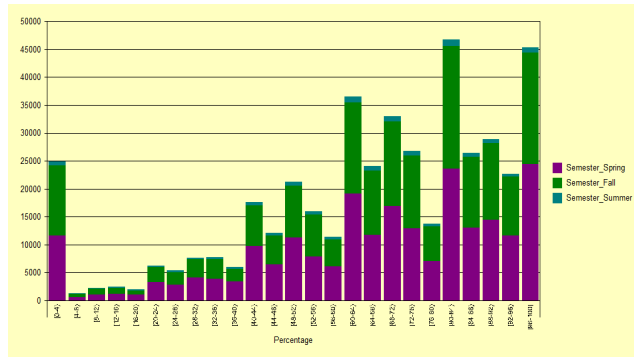

*Figure 8b: Performance of students by distribution of percentage across semesters*

In figure 8b, the student performance is analyzed based on student ratio. In this research work, rapid changes are found in *semester_spring* and *semester_fall* while compared with *semester_summer*. This is because of the student who are all belongs to *semester_summer* have less number of laboratory mark compared to both *semester_spring* and *semester_fall*. In this scenario, the system chooses the attribute '*Semester & Percentage of marks*' and performed cluster analysis with the help of software agents. Further mining of this results are provided in figure 8c.

International Journal of Artificial Intelligence & Applications (IJAIA), Vol.3, No.1, January 2012

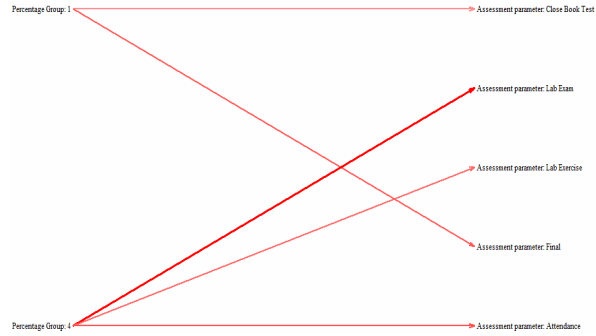

*Figure 8c: Link chart represents the relation between Percentage and Assessment Component (Correlation Value: 1.03e+003 and only positive links)*

In figure 8c, the multidimensional visualization is one of the major events in this research work with respect to user constrain for better understanding and meaningful knowledge discovery in different manner so that, the domain knowledge is not all necessary to analyze the visual reports. The visualization is based on the data types so that, the decision agent will select the appropriate visual tool with respect to dimensionality. ie 1D, 2D, 3D and etc such that, the clusters after cluster quality detection are made visualize by means of decision agent. The chart shown is one of the results after cluster detected by the user, in which the thick red line shows stronger cluster in which more attributes are involved to produce the final result which is also meaning full and easy to identify knowledge. These processes are carried out by intelligent agent in the name of decision agent with lesser time when compared to manual process. This is one the major advantage while compared with other frame work. The chart shown here is link chart in which different attributed are linked together to identify group of students, who are all secured more mark with respect to different parameters based on assessments. In this scenario, the system chooses the attribute '*Semester, Percentage of marks & Assessment parameter*' and performed cluster analysis with the help of software agents. Further mining of this results are provided in figure 8d.

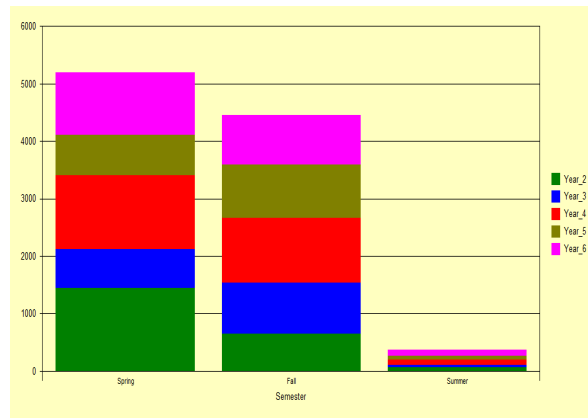

*Figure 8d: Group4 students (Percentage greater than 90) performance in Lab Exam across semesters*

From the above figure 8d, analysis takes place on semester wise on overall percentage of *marks vs total number of students* and *years*. According to this research work, students who belong to *spring_semester* and *fall_semester* obtain more marks in the entire subject with corresponding year. As per our research work, this much deviation is due to laboratory marks are also counted in both *spring_semester* and *fall_semester* but not in *summer_semester*.

135

International Journal of Artificial Intelligence & Applications (IJAIA), Vol.3, No.1, January 2012## 9. CONCLUSION

Automated data mining is an emerging concept in data processing. In this research work, entire mining process is carried out by different intelligent agents with the help of automated system based on the user history. The major aim is to satisfy the user expectation and interest. This research work mainly focused on automated mining process from which the suitable mining techniques are analyzed based on the data set and attribute nature. Clusters are formulated and also a detection technique is used to analyze the quality of the cluster with respect to data base selected by the user. Thus, intelligent agent identifies whether new cluster is of good quality or not based on the user navigation, which is referred as cluster detection for visualization. Visualization is based on the nature of the elements within the selected cluster. This is because; clusters will never give meaning results because of non domain experts when compared with domain experts. The domain experts will have more knowledge and easily understand the system within a short span of time. The solutions discussed in this research work is more scalable and compatible when compared to the present existing systems by means of less attentions and more user friendly approach to a focused results.

## REFERENCES

[1] Ayse Yasemin SEYDIM "Intelligent Agents: A Data Mining Perspective" Southern Methodist University, Dallas, 1999.

[2] Abou-Rjeili, G. Karypis, "Multilevel Algorithms for Partitioning Power-Law Graphs," IEEE International Parallel & Distributed Processing Symposium (IPDPS), 2006.

[3] A. Bertoni and G. Valentini. "Discovering structures through the Bernstein inequality". In KES-WIRN 2007, Vietri sul Mare, Italy, 2007.

[4] Berry, M. and Linoff, G" Mastering Data Mining: The Art and Science of customer Relationship Management", Wiley, 1999.

[5] Charlie Y. Shim, Jung Yeop Kim, Constructing Cost-effective Anomaly Detection mputer Science 2008, WCECS 2008, October 22-24,2008, San Francisco, USA.

[6] D. Dasgupta, F. Gonzalez, K. Yallapu, J. Gomez, R. Yarramettii, CIDS: An Agent- Based Intrusion Detection System, Computers & Security pp 387-398, 2005.

[7] Don Gilbert, "Intelligent Agents: The Right Information at the Right Time" IBM Corporation, Research Triangle Park, NC USA, May, 1997.

[8] T.Dean, J.Allen, Y.Aloimonos, "Artificial Intelligence: Theory and Practice", The Benjamin/Cummings Publishing Co. Inc., 1995.

[9] Eleni Mangina, Intelligent Agent-Based Monitoring Platform for Applications in Engineering, International Journal of Computer Science & applications Vol.2, No.1, pp. 38-48, 2005.

[10] Elth Ogston, Benno Overinder, Maarten van Steen, and Frances Brazier, A Method for Decentralized Clustering in Large Multi-Agent Systems, AAMAS'03, July 14-18, 2003, Melbourne, Australia, ACM 2003.

[11] M. Goebel, L. Grunewald , "A Survey of Data Mining and Knowledge Discovery Software Tools", SIGKDD Explorations, Vol 1, pp. 20-33, 1999.

[12] J. Han and M. Kamber. "Data Mining, Concepts and Technique". Morgan Kaufmann, San Francisco, 2001.

[13] Henrik et al., "Methods for Visual Mining of Data in Virtual Reality", Proceedings of The International Workshop on Visual Data Mining (VDM@PKDD2001), 2001.

[14] Hing-Yan Lee, Hwee-Leng, Eng-whatt Toh. "A multi-Dimensional Data Visualization Tools for Knowledge Discovery in Databases", IEEE 0730-3157/1995.136

International Journal of Artificial Intelligence & Applications (IJAIA), Vol.3, No.1, January 2012

International Journal of Artificial Intelligence & Applications (IJAIA), Vol.3, No.1, January 2012

**AUTHOR PROFILES:**

Jayabrabu R obtained Master Degree in Computer Science and Application from Pondicherry University, Pondicherry, India in 2004 and Master of Philosophy in Annamalai University, Chidhambaram, Tamil Nadu and India in 2007.  He is a research scholar of Bharathiar University, Coimbatore, India, he is working as Associate Professor.   His Areas of Interests are in Intelligent Agents, Data Analysis, and Clustering.

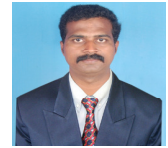

Dr. V. Saravanan obtained Ph.D in Computer Science in 2004 from Bharathiar University, Coimbatore, Tamil Nadu, India.  He has produced 3 Ph.D Candidates.  He has published articles in 15 International Journals.  His area of interest includes Data Mining and Software Agents.

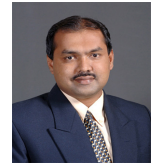

Dr.K.Vivekanandan obtained Ph.D in Computer Science in 1996. He has produced 10 Ph.D candidates and 7 M.Phil candidates. He has published articles in 17 International Journals. His area of interest includes Data Mining and Information Systems.

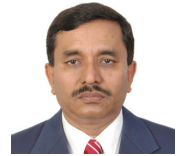